\documentclass[letterpaper, 10 pt, conference]{ieeeconf}
\usepackage{amsmath}
\usepackage{graphicx}
\usepackage{subfigure} 
\usepackage{esvect}
\usepackage{gensymb}
\usepackage{listings}
\usepackage{xcolor}
\usepackage{mathtools}
\usepackage{etoolbox, xparse}
\usepackage{amssymb}
\usepackage{booktabs}
\usepackage{multirow}
\usepackage{balance}
\usepackage{xpatch}
\makeatletter
\patchcmd\@makecaption{\\}{.~}{}{\fail}
\makeatletter

\makeatletter
\newif\if@restonecol
\makeatother

\usepackage[ruled,vlined,commentsnumbered, lined, boxed, shortend, onelanguage,nokwfunc]{algorithm2e}
\SetKwInput{Parameter}{Parameter}
\SetKwInput{Operation}{Operation}
\SetKwRepeat{Do}{do}{while}
\SetKw{Continue}{continue}
\SetKw{break}{break}
\usepackage{algpseudocode}
\usepackage{amsmath}
\definecolor{codegreen}{rgb}{0,0.6,0}
\definecolor{codegray}{rgb}{0.5,0.5,0.5}
\definecolor{codepurple}{rgb}{0.58,0,0.82}
\definecolor{backcolour}{rgb}{0.95,0.95,0.92}

\lstdefinestyle{mystyle}{
	backgroundcolor=\color{backcolour},
	commentstyle=\color{codegreen},
	keywordstyle=\color{magenta},
	numberstyle=\tiny\color{codegray},
	stringstyle=\color{codepurple},
	basicstyle=\ttfamily\footnotesize,
	breakatwhitespace=false,
	breaklines=true,
	captionpos=b,
	keepspaces=true,
	numbers=left,
	numbersep=5pt,
	showspaces=false,
	showstringspaces=false,
	showtabs=false,
	tabsize=2
}

\IEEEoverridecommandlockouts

\title{\LARGE \bf
Barycode-based GJK Algorithm
}

\author{Yu Zhang$^{1}$, Yangming Wu$^{1}$, Xigui Wang$^{1}$, Xiaocheng Zhou$^{1}$
\thanks{$^{1}$Yu Zhang, Yangming Wu, Xigui Wang and Xiaocheng Zhou are with UISEE Technology Co. Ltd. \{
{\tt\small yu.zhang.bit@gmail.com};
{\tt\small yangming.wu@uisee.com};
{\tt\small xigui.wang@uisee.com};
{\tt\small xiaocheng.zhou@uisee.com} \}}%
}

\begin{document}

\maketitle
\thispagestyle{empty}
\pagestyle{empty}

%%%%%%%%%%%%%%%%%%%%%%%%%%%%%%%%%%%%%%%%%%%%%%%%%%%%%%%%%%%%%%%%%%%%%%%%%%%%%%%%
\begin{abstract}
In this paper, we present a more efficient GJK algorithm to solve the collision detection and distance query problems in 2D.  We contribute in two aspects:
First, we propose a new barycode-based \emph{subdistance} algorithm that does not only provide a simple and unified condition to determine the minimum simplex but also improve the efficiency in distant, touching and overlap cases in distance query. Second, we provide a highly efficient implementation subroutine for collision detection by optimizing the exit conditions of our GJK distance algorithm, which shows dramatic improvements in run-time for applications that only need binary results. We benchmark our methods along with that of the well-known open-source collision detection libraries, such as Bullet, FCL, OpenGJK, Box2D, and Apollo over a range of random datasets. The results indicate that our methods and implementations outperform the state-of-the-art in both collision detection and distance query.
\end{abstract}

%%%%%%%%%%%%%%%%%%%%%%%%%%%%%%%%%%%%%%%%%%%%%%%%%%%%%%%%%%%%%%%%%%%%%%%%%%%%%%%%
\section{INTRODUCTION}
The problem of collision detection and distance computation has been the subject of various research fields including robotics\cite{pan2012fcl, pan2013real}, computational geometry\cite{cgal}, simulation\cite{bullet}, gaming engine\cite{Box2D}. It is one of the most time-consuming process of motion planning subsystem for self-driving, especially in sampling-based motion planning framework. It is also the essential module that guarantees the safety of self-driving system. Different from that of applications in simulation and game development, more strict requirements for collision detection module have been raised in self-driving system in terms of  efficiency, robustness and richness of information it provides. In this paper, we mainly focus on the efficiency of the collision detection.

%\textbf{Richness of the Information}
%Collision detection or overlapping test is usually used in sampling-based motion planning framework to determine the feasibility of the samples. In random sampling-based frame work, the samples are generated randomly, thus feasibility check by collision detection is enough. But in deterministic sampling-based motion planning framework, further information such as witness points and the closest direction or the fastest separate direction can be used to adjust the sampling strategies to improve the sampling efficiency.
%
%Trajectory optimization methods such as Chomp \cite{zucker2013chomp}, Trajopt \cite{trajopt2014ijrr}, CFS \cite{liu2018convex} need to calculate the signed distance and distance gradient to find the update direction of the decision variables in mathematical optimization. Detecting intersection to provide a binary result or providing a distance without any direction information are not enough any more. The distance, witness points and direction (gradient) information can be leveraged to improve the performance and quality of the planning result in more complex situation.
%
%
%For decision-making and high-level motion planning algorithms, the semantic information of the collision status can be used to analyse the risk degree of a behavior. In order to improve the safety and feasibility of decision-making, semantic information such as a collision point in a vehicle body, crash direction of an  obstacle vehicle, or the severity of a traffic accident, can be provided for the self-driving system.

It's well-known that a bounding volume hierarchy (BVH) or space-partition-based collision detection system is able to leverage certain data structure to reduce unnecessary times of the narrow phase collision detection algorithms, but this kind of framework do little improvements for object-pairs that are all actually collide. Thus, improving the efficiency of the narrow phase collision detection algorithms is still essential.

Many narrow phase algorithms have been developed to solve the distance query problem over the decades, including the separating axis theorem (SAT) algorithm \cite{gottschalk1996separating}, the Lin and Canny algirithm \cite{lin1991fast}, Voronoi-clip (V-clip) algorithm \cite{mirtich1998v}, the Gilbert-Johnson-Keerthi (GJK) algorithm \cite{bergen1999fast, cameron1997enhancing, montanari2017improving}.
Among them, the GJK algorithm is the most versatile algorithm to solve above problems that raises the least requirements for the geometry representations of objects while has the comparable computational efficiency and the strong extendibility in the same framework\cite{montanari2017improving}. For example, the penetration depth calculation (with the help of the EPA algorithm), raycast or shapecast functionality can be developed and extended in the same framework with ease. Any convex shape that is able to calculate the support mapping in a given direction can be handled in the framework.

As we all know, the general implementation of collision detection algorithms in narrow phase are usually written in 3D for games \cite{Box2D}, physics engine\cite{todorov2012mujoco}, robotics\cite{pan2012fcl}. For autonomous driving, a large amount of collision detection in motion planning and decision-making system only need to be taken in 2D space. Collision detection may be executed million times per planning loop in complex scenarios. Applying general 3D collision detection algorithms brings noticeable unnecessary runtime overheads to the system, especially when the system is running on embedded devices. This can be proved in our benchmarks on random datasets.

In this paper, we present a more efficient GJK algorithm to solve the collision detection and distance query problems in 2D.  We contribute in two aspects:
First, we propose a new barycode-based \emph{subdistance} algorithm that does not only provide a simple and unified \label{key}condition to determine the minimum simplex but also improve the efficiency in distant, touching and overlap cases in distance query. Second, we provide a highly efficient implementation routine for collision detection by optimizing the exit conditions of our GJK distance algorithm, which shows dramatic improvements on run-time for applications that only need binary results. 

The structure of the paper is as follows. The methodology is presented in Section II. Section III shows the benchmark setup and results. Section IV draws the conclusion.

%2. efficient, early return of SAT may be faster in separate cases. But our early exit condition in collision detection routine make the GJK a better solution overall.

%\begin{itemize}
%	\item Johnson
%	\item Johnson-backup
%	\item Voronoi
%	\item Signed Volume
%\end{itemize}

%Existing issues:
%\begin{itemize}
%	\item Numerical error
%	\item Robustness: degenerated cases;local maximum
%\end{itemize}

\section{GJK}
\subsection{Problem Formulation}
The problem of collision detection and distance query can be summarized as follow:

Given two objects $P, Q \in R^n$ represented by non-empty and finite convex sets of points $x$ in $R^n$, find the minimum Euclidean distance $\text{dist}(P, Q)$ determined by two witness points
$p \in P$ and $q \in Q$ such that:
\begin{equation} \label{eq:distance_vec}
\text{dist}(P, Q) = \lVert p  - q \rVert = \text{min} \big\{ \lVert x - y \rVert \mid x \in P, y \in Q \big\}
\end{equation}

For collision detection problem, only the sign of the distance $\text{dist}(P, Q)$ matters in a way that the positive sign means collision-free, otherwise in collision. For the distance problem, only the scalar of the $\text{dist}(P, Q)$ is needed. Most of the applications leverage only the scalar of the separating vector $\text{dist}(P, Q)$ to check the collision status and evaluate collision risks in distant cases.
For us, the complete distance problem is to find not only the minimum distance $\text{dist}(P, Q)$ but also witness points $p, q$ and its separating vector $\boldsymbol{\vec{pq}}$ between two convex sets.

It is worth noting that the separating vector $\boldsymbol{\vec{pq}}$ is emphasized as one of the requirements for a complete distance problem.  It's mainly motivated by two aspects. First, for $\text{dist}(P, Q) = 0$, the separating direction can not be acquired only from the witness points. Additionally, when the two witness points are too close, the calculation of the direction formed by witness points will face numerical issues. Second, for the point contact cases, there exists a cone region instead of a single separating direction. Thus the separating direction has to be calculated in the individual algorithm somehow and output.

To be clear, we divide the requirements for the problem of collision detection and distance query to three levels:
\begin{itemize}
	\item Level 1: binary result, collide or not
	\item Level 2: distance scalar
	\item Level 3: distance, complete witness points and separating direction
\end{itemize}
Due to the different requirements for solutions of the problem, implementation varies and may affect the runtime efficiency dramatically. These requirements can be also used to identify the capacity of the underlying implementations and exploited to construct the most efficient algorithm.
Above statements apply to both two and three dimensional problems. Due the limited space, we focus on only Level 1 and Level 2 in the $R^2$ Euclidean space in this paper. And the points $x$ of convex shape $P, Q$ are ordered counter-clock-wise. For readers who are interested in Level 3 requirements, please refer to our following long version of this paper.

\subsection{How GJK works}

%Due to numerical and rounding errors, the
%detection result may be false in two ways:
%false negative, when the algorithm fails to detection the collision event; false positive, when the algorithm reports collision while the real ones do not happen. from [Defending Continuous Collision Detection against Errors]

The GJK is a decent iterative algorithm that leverages the support mapping, configuration space obstacle (CSO) and simplex concepts to determine the minimum distance between two convex sets.

\textbf{Configuration space obstacle}
Recall the distance vector between two convex sets in \eqref{eq:distance_vec} in the problem formulation. The representation of the distance vector actually matches the definition of the \textit{configuration space obstacle} (CSO), which is the Minkowski difference between two convex sets as follow:
\begin{equation}
	\text{CSO} = P \ominus Q = \big\{x - y \mid x \in P, y \in Q \big\}
\end{equation}
where $\ominus$ is the Minkowski difference operator. The problem of finding the minimum distance of $\lVert x - y \rVert$ mathematically equals to that of finding the point $z_{CSO}$ in $\text{CSO}$ such that
\begin{equation}
	\lVert z_{CSO} - O \rVert = \text{min} \lVert v - O \rVert, v \in \text{CSO}
\end{equation}
The $z_{CSO}$ may lie in the vertex or the edge of the CSO. And $z_{CSO}$ is also called the closest point of the origin $O$ on the CSO in this paper.
Since the Minkowski difference of two convex set is convex, the CSO is convex. There exists an unique solution to the minimum norm problem of the distance vector from the origin $O$ to the point $z_{CSO}$. This elegant transformation of the problem not only makes the distance calculation procedure simplified but also makes the solution more expressive. By expressive, we mean the unique solution $z_{\text{CSO}}$ can represent infinite pair of witness points($p, q$) for polygons with parallel edges ( polytopes with parallel faces in 3D). We can also convert the unique solution in CSO space to solutions in actual physical space with the help of simplex and barycentric coordinates $\lambda$.
Now the problem becomes whether $\text{CSO} \cap \text{O} = \emptyset$. If true, the corresponding minimal distance vector between
the origin O and CSO need to be found. Otherwise, two shapes collide and the zero distance is returned. 

\textbf{Support Mapping}
Calculating the full set of CSO to find the solution is time-consuming. In fact, we only need a subset of CSO that is close to the origin $O$ instead of the full explicit $\text{CSO}$. For efficiency, we only need the minimal subset that is also able to represent the closest point $z_{\text{CSO}}$. Then we say all the vertices of the minimal subset supports $z_{\text{CSO}}$. Thanks to the support mapping and convex hull property of the $\text{CSO}$, we can find the furthest vertex in $\text{CSO}$ in a given direction. The support function of the support mapping in a specified direction $\boldsymbol{\vec{v}_s}$ is defined as
\begin{equation}
   \text{S}_{\text{CSO}}(\boldsymbol{\vec{v}_s}) = 
   \text{S}_{\text{P} \ominus \text{Q}}(\boldsymbol{\vec{v}_s}) = 
   \text{max}\{ v \cdot \boldsymbol{\vec{v}_s} \mid v \in \text{CSO} \}
\end{equation}
Intuitively, the support mapping could move the subset of CSO towards the origin by adding a new vertex that is closer to the origin. And we only care about the vertices that contribute to the calculation of the closest point $z_{\text{CSO}}$. This is where the simplex concept becomes useful.

\textbf{Simplex}
A $n$-simplex is a convex hull of $n+1$
affinely independent points $z_i$ \cite{ericson2004real}. More formally, the simplex can be represented by 
\begin{equation}
\begin{aligned}
	conv(Z) =  \{&\lambda_0z_0 + \cdots + \lambda_nz_n \lVert \\
	&\sum_{i= 0}^{n}\lambda_i = 1 \ \text{and} \  \lambda_i \geq 0 \ \text{for all} \  i\}
\end{aligned}
\end{equation}
where $Z = \{z_i\}, i = 0, \cdots, n$.
 According to the Caratheodory's theorem, a point $z \in conv(Z) \subset R^n$ can be expressed by a convex combination of n+1 points. And fewer points can be used for some special cases, for example, the closest points on the edge or vertex in 2D. Thus during the iteration, we only need to maintain at most n+1 vertices generated by supporting mapping (n = 2 in 2D, n = 3 in 3D). And the closest point $z_{\text{s}}$ of the origin O onto the simplex can be expressed by the barycentric coordinates with the corresponding simplex. In this way, we can track the distance vector $\boldsymbol{v}$ defined by the closest point $z_\text{s}$ and the origin.

With the three core concepts used by GJK established, the GJK main loop can be explained as below(shown in Alg. \ref{alg:gjk}). 

Starting with an heuristic support direction $v_{\text{init}}$ and an empty simplex set $W_k$, the GJK algorithm generates the start CSO point $v_k$ in $v_{\text{init}}$ support direction. Then the algorithm begins the iteration by searching the $-v_k$ direction and generating the support point $w_k$. If $w_k$ is not further than the CSO vertex $v_k$ in $-v_k$ direction, the algorithm terminates and return the $v_k$ as the final distance solution. Otherwise, the $w_k$ is added to the simplex set $W_k$ to form the new simplex set $\tau_k$. The \emph{sub-distance} algorithm is responsible for finding the minimum convex subset $W_k$ of $\tau_k$ that still contains the new closest point $v_{k}$. The new $-v_{k}$ will be the support direction for the next iteration if the algorithm continues. The algorithm terminates until the region defined by $W_k$ encloses the origin $O$ or the norm $\lVert v \rVert$ can not be reduced within a given tolerance for distance. 

\begin{algorithm} \label{alg:gjk}
	\caption{GJK Distance Query Algorithm (2D, adapted from \cite{montanari2017improving})}
	\KwIn{Convex  Shape:${P, Q}$}
	\KwOut{Distance:${\lVert v_{k+1} \rVert}$}
	\SetKwFunction{S}{S}
	\SetKwFunction{subdistance}{subdistance}
	\SetKwFunction{card}{card}
	\Operation{\card cardinality of a set}
	\Parameter{$k_{\text{max}}$ max iteration number allowed}
	\Parameter{$\varepsilon \in R_{++}$ the positive tolerance}
	\If{$\big(\exists v \in (\text{P} \ominus \text{Q})\big) \land (\lVert v \rVert \neq 0)$}{$v_{\text{init}}$ = $v$}
	\Else{
		\Return 0\;
	}
	$v_1$ = $\S_{\text{P} \ominus \text{Q}}(-v_{\text{init}})$\;
	$\tau_k = \{v_1\}$, $\text{W}_k = \{v_1\}$, $k = 0$ \;
	\Do{ $ \big(\card(W_k) < 3 \big) \land (k < k_{\text{max}} )$}
	{
		k = k+1,
		$w_k$ = $\S_{\text{P} \ominus \text{Q}}(-v_k)$\;
		\If{$\lVert v_k \rVert^2 - v_k \cdot w_k \leq  \varepsilon^2 \lVert v_k \rVert^2$} {
			$v_{k+1} = v_{k}$\;
			\break\;
		}
		
		$\tau_k = {w_k} \cup W_{k-1}$\;
		[$W_k, \lambda, v_{k+1}$] = \subdistance($\tau_k$)\;
		\If{$\lVert v_{k+1} \rVert \leq \varepsilon $} {
			\break\;
		}
	}
	\Return $\lVert v_{k+1} \rVert$\;
\end{algorithm}

The main GJK loop for all the variants are basically the same. And the major differences of GJK variants lie in the \emph{subdistance} algoirthms. The GJK algorithms in Bullet \cite{bullet}, FCL \cite{pan2012fcl}, libccd \cite{libccd}  and Box2D \cite{Box2D} libraries are all voronoi-region-based methods to find the best convex subset and the colsest points with the minimum distance. They exploit a bottom-up manner to search all possible voronoi regions to find the minimal distance vector \cite{ericson2004real}. Among them, the GJK of Box2D is the most efficient implementation in 2D according to our benchmark. The latest work about GJK \cite{montanari2017improving} introduces a signed-volume(SV)-based method in sub-distance algorithm that is able to improve the numerical robustness while reducing the computational time at the same time. The SV-based sub-distance algorithm leverages the fact the barycentric coordinates are invariant to affine transformations to solve a simpler linear system. Different from the former variants, it search the minimum norm in a top-down manner, which resulting a shorter search path.

\subsection{Barycode-based subdistance algorithm}
We proposed a new barycode-based (short for barycentric coordinates code) \emph{subdistance} algorithm that does not only provide a simple and unified \label{key}condition to determine the minimum convex subset of $\tau_k$ but also improve the efficiency in distant, touching and overlap distance calculation cases. The proposed method also employs a top-down manner to search the solution and leads to an even shorter search path than the SV-based method with the help of the barycode and voronoi region. 

The general \emph{subdistance} algorithm is as below:
\begin{algorithm} \label{alg:subdist}
	\caption{barycode-based \emph{subdistance} algorithm}
	\KwIn{$\tau_k$ new simplex set}
	\KwOut{$W_k$ minimal simplex set,\\
	       $\lambda$ barycentric coordinats,\\
           $v_{k+1}$ the minimum distance vector}
	\SetKwFunction{S}{S}
	\SetKwFunction{SubDistance}{SubDistance}
	\SetKwFunction{card}{card}
	
	\Operation{\card cardinality of a set}
	n = \card($\tau_k$)\;
	\Switch{n} {
		\Case{3} {
		[$W_k, \lambda, v_{k+1}$] = S2D($\tau_k$),
		\break\;
		}
		\Case{2} {
		[$W_k, \lambda, v_{k+1}$] = S1D($\tau_k$),
		\break;
		}
		\Other{
		$W_k = \tau_k , \lambda = \{1.0\}, v_{k+1} = v \in W_k$,
		\break\;
		}
	}
	\Return [$W_k, \lambda, v_{k+1}$]\;
\end{algorithm}
By calculating the cardinality of the simplex set $\tau_k$, that is the number of elements in $\tau_k$, the minimum subdistance problem recasts to the problem of the minimum distance from the origin $O$ to a vertex, a line segment (S1D problem) or a triangle (S2D problem) in CSO.
For the S1D problem, we use the dot product of two vectors to identify the voronoi regions of a line segment and reuse the conditions to calculate the barycentric coordinates if the closest point lie on the line segment, which is shown in algorithm \ref{alg:s1d}.
This leads to a slight faster implementation than that of 2D version of SV-based S1D routine.

\begin{algorithm} \label{alg:s2d}
	\caption{S2D($\tau_k$)}
	\KwIn{$\tau_k = \{A, B, C\}$ the simplex set}
	\KwOut{$W_k$ the minimal simplex set,\\
		$\lambda$ barycentric coordinats,\\
		$v_{k+1}$ the minimum distance vector}
	\SetKwFunction{SameSign}{SSign}
	\SetKwFunction{ConeRegion}{ConeRegion}
	\SetKwFunction{FMain}{SSign}
	\SetKwProg{Fn}{Function}{:}{}
	\Fn{\FMain{$m, n$}}{
		\KwRet ($(m > 0) == (n > 0)$)\;
	}
%	$A = s_3, B = s_2, C = s_1$\;
	$\sigma_u = B \times C, \sigma_v = C \times A, \sigma_w = A \times B$\;
	$sum = \sigma_u + \sigma_v + \sigma_w$\;
	$
	\begin{aligned}
	barycode = &\SameSign(sum, \sigma_w) \lor \big(\SameSign(sum, \sigma_v) \ll 1\big)\\
	& \lor (\SameSign(sum, \sigma_u) \ll 2)
	\end{aligned}
	$\;
	\Switch{barycode} {
		\Case{1} {
			[$W_k, \lambda, v_{k+1}$] = \ConeRegion($\tau_k, C$), \break\;
		}
		\Case{2} {
			[$W_k, \lambda, v_{k+1}$] = \ConeRegion($\tau_k, B$), \break;
		}
		\Case{3} {
			$
			\begin{aligned}
			\tau_k = \tau_k \setminus \{A\}, [W_k, \lambda, v_{k+1}] = \text{S1D}(\tau_k)
			\end{aligned}
			$, \break;
		}
		\Case{4} {
			[$W_k, \lambda, v_{k+1}$] = \ConeRegion($\tau_k, A$), \break;
		}
		\Case{5} {
			$
			\begin{aligned}
			\tau_k = \tau_k \setminus \{B\}, [W_k, \lambda, v_{k+1}] = \text{S1D}(\tau_k)
			\end{aligned}
			$, \break;
		}
		\Case{6} {
			$
			\begin{aligned}
			\tau_k = \tau_k \setminus \{C\}, [W_k, \lambda, v_{k+1}] = \text{S1D}(\tau_k)
			\end{aligned}
			$, \break;
		}
		\Case{7} {
			$
			\begin{aligned}
			&\lambda_u = \frac{\sigma_u}{sum}, \lambda_v = \frac{\sigma_v}{sum}, \lambda_w = 1 - \lambda_u - \lambda_v\\
			&\lambda = \{\lambda_u, \lambda_v, \lambda_w\}, W_k = \tau_k\\
			&v_{k+1} = \lambda_u \cdot A + \lambda_v \cdot B + \lambda_w \cdot C
			\end{aligned}
			$\;
			\break;
		}
	}
	\Return [$W_k, \lambda, v_{k+1}$]\;
\end{algorithm}

\noindent \textbf{S2D subroutine and Barycentric Coordinates}

%\begin{figure}[h!]
%	\centering
%	\includegraphics[width=0.3\linewidth]{images/barycentric_region.png}
%	\caption{Barycentric Regions of A 2-D simplex}
%	\label{fig:baryregion}
%\end{figure}

\begin{figure}
	\centering
	\subfigure[Barycentric regions]{
		\label{fig:baryregion} %% label for first subfigure
		\includegraphics[width=1.4in]{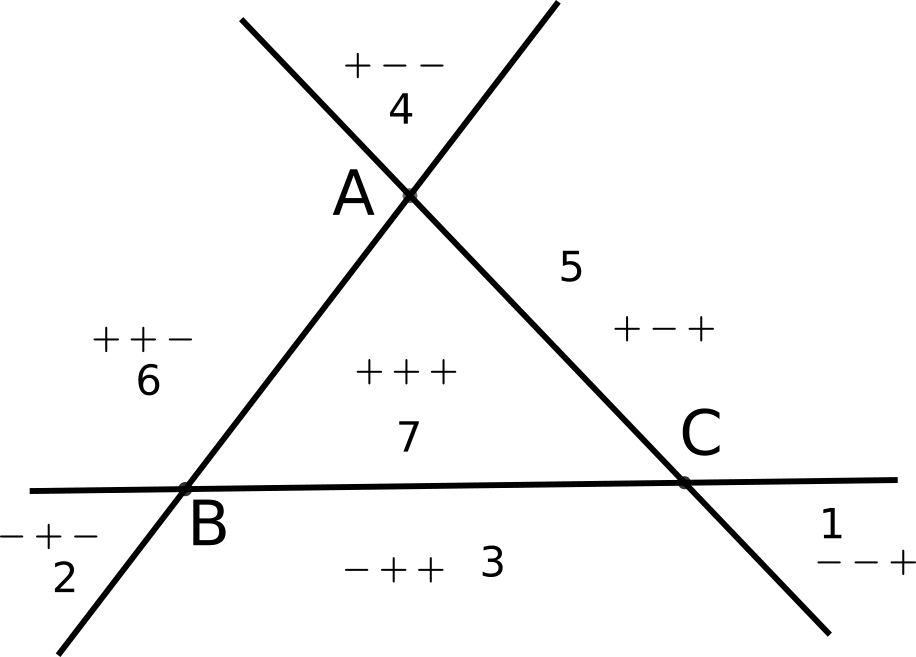}
		}
	\hspace{0.05in}
	\subfigure[Voronoi regions]{
		\label{fig:voronoi} %% label for second subfigure
		\includegraphics[width=1.4in]{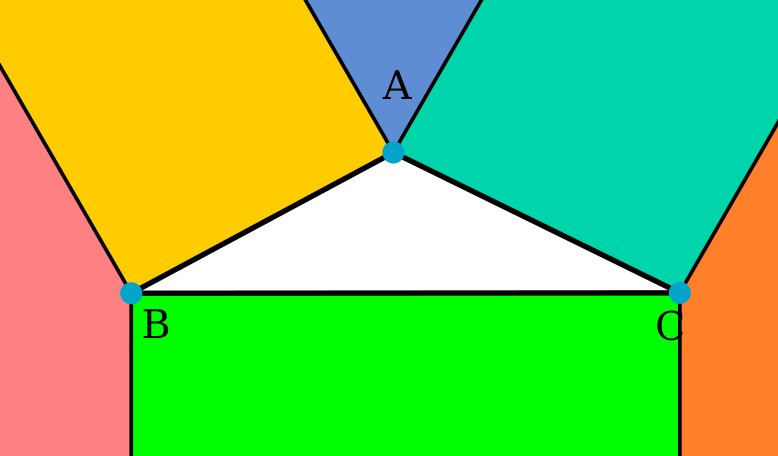}
		}
	\caption{Different regions for S2D subroutine}
	\label{fig:s2dregion} %% label for entire figure
\end{figure}

Assume $P$ is an arbitrary point in the plane, the cartisian coordinates of the $P$ can be expressed by the barycentric coordinates $(u, v, w)$ that corresponds to three non-collinear points $A, B, C$ respectively as below:
\begin{equation}
\begin{aligned}
&P = u \cdot A + v \cdot B + w \cdot C,\ \ u + v + w = 1.0 \\
%&P = (1 - v - w)\cdot A + v \cdot B + w \cdot C \\
%&v \cdot (B - A) + w \cdot (C - A) = P - A
\end{aligned}
\end{equation}

In general, the barycentric coordinates
can be calculated by solving a $2 \times 2$ linear system \eqref{eq:linear} as below with Cramer's rule according to \cite{ericson2004real}.
\begin{equation}\label{eq:linear}
  \begin{aligned}
  (v \cdot \overrightarrow{AB} + w \cdot \overrightarrow{AC}) \cdot \overrightarrow{AB} = \overrightarrow{AP} \cdot \overrightarrow{AB}\\
  (v \cdot \overrightarrow{AB} + w \cdot \overrightarrow{AC}) \cdot \overrightarrow{AC} = \overrightarrow{AP} \cdot \overrightarrow{AC}
  \end{aligned}
\end{equation}

\begin{equation}
\begin{aligned}
	&denom = \lVert \overrightarrow{AB} \times \overrightarrow{AC} \rVert\\
	&v = \frac{\lVert \overrightarrow{AP} \times \overrightarrow{AC} \rVert}{denom}, 
	w = \frac{\lVert \overrightarrow{AB} \times \overrightarrow{AP} \rVert}{denom},
	u = 1 - v - w
\end{aligned}
\end{equation}
A nice property of the barycentric coordinates is that the sign of the coordinate triplet $(u, v, w)$  divides the plane to 7 unique regions, shown as the Fig. \ref{fig:baryregion}. To make it easy to represent, we use a 3-bit number to encode the region, which is called the \textit{barycode} in this paper. The signs of $u, v, w$ correspond to the \textit{barycode} bit from the high bit to the low bit respectively (line 5 in Alg. \ref{alg:s2d}). A positive sign means 1 and otherwise 0 for the corresponding bit. This leads to an unique $barycode$ for every individual region of the plane. Additionally, we can determine the $barycode$ with the intermediate results of the calculation of the barycentric coordinates and reuse the intermediate results to get the exact values of barycentric coordinates  later when they are actually needed. The $barycode$ is able to quickly identify the minimum convex subset of $\tau_k$ that contains $v_{k+1}$. 
For example, the barycode 3, 5, 6 refer to $\{B,C\}, \{A,C\}, \{A,B\}$ subsets respectively. Then the S1D subroutine can be used to do the final solving. The barycode 7 means the point P is inside the triangle, which means a collision. Then the exact barycentric coordinates can be calculated by the line 20 of the Alg. \ref{alg:s2d}. The barycode 1, 2, 4 refer to the C, B, A vertex cone region. This kind of region is the most complicated case of the whole algorithm. The solving procedure is defined by the algorithm \ref{alg:coneregion} ConeRegion. It firstly checks whether the angle of the cone region defined by the $barycode$ is acute. 
\begin{equation}
\begin{aligned}
\vv{\text{MV}} \cdot \vv{\text{NV}} \ge 0,\ &\text{if} \ \angle{\text{MVN}} \le 90 \degree \\
\vv{\text{MV}} \cdot \vv{\text{NV}} < 0, \ &\text{if} \ \angle{\text{MVN}} > 90 \degree
\end{aligned}
\end{equation}
If true, the minimum subset $W_k = \{ V\}$ is clearly the vertex $V$. Otherwise, we need further conditions to determine the solution. The SV-based GJK algorithm handles this region by calculating all the possible solutions, comparing them and outputting the one with the minimum norm. We instead leverage the orthogonal condition (line 3, 5 in Alg. \ref{alg:coneregion}) to determine which direction to progress. Then it exploits the S1D subroutine to solve the problem. This leads to a shorter search path. If the origin O is not in the voronoi regions of VM and VN, it falls into the voronoi region of the vertex V, which is automatically merged to the acute angle case (line 7 in Alg. \ref{alg:coneregion}). 

The voronoi-based GJK \cite{Box2D} is based on 7 voronoi regions (shown in Fig. \ref{fig:voronoi}) to determine the supporting vertices for the triangle simplex cases. Unlike our barycode-based GJK, its conditions to determine the individual region to project are all different. Thus in the worst case, the algorithm has to go through 7 different cases to get the minimum distance vector. For example, the implementation of Box2D checks the case that encloses the origin at the end, thus it need more time to detection the overlap comparing to distant cases. Our algorithm overcomes this flaw and keep about the same runtime efficiency on average for distant, touching and overlap cases. 

%\begin{figure}[h!]
%	\centering
%	\includegraphics[width=0.8\linewidth]{images/voronoi_region.png}
%	\caption{Voronoi region for S2D}
%	\label{fig:voronoi}
%\end{figure}
\begin{algorithm} \label{alg:coneregion}
	\caption{ConeRegion($\tau_k, V$)}
	\KwIn{$\tau_k = \{ s_1, s_2, s_3\}$ the new simplex set, \\
		$V$ the vertex of the cone region}
	\KwOut{$W_k$ the minimal simplex set,\\
		$\lambda$ barycentric coordinats,\\
		$v_{k+1}$ the minimum distance vector}
	\SetKwFunction{S}{S}
	\SetKwFunction{SubDistance}{SubDistance}
	$\{ M, N\} = \tau_k \setminus \{V\}$\;
	\If{$\overrightarrow{MV} \cdot \overrightarrow{NV} < 0 $}{
		\If{$\overrightarrow{OV} \cdot \overrightarrow{MV} > 0$}{
						$\tau_k = \{V, M\}$\;
						$[W_k, \lambda, v_{k+1}] = \text{S1D}(\tau_k)$\;
			\Return $[W_k, \lambda, v_{k+1}]$ \;
%			\Return ${ \text{S1D}(\tau_k = \{V, M\})}$\;
		}
		\If{$\overrightarrow{OV} \cdot \overrightarrow{NV} > 0$} {
						$\tau_k = \{V, N\}$\;
						$[W_k, \lambda, v_{k+1}] = \text{S1D}(\tau_k)$\;
						\Return $[W_k, \lambda, v_{k+1}]$ \;
%			\Return ${ \text{S1D}(\tau_k = \{V, N\})}$\;
		}
	}
		$W_k = \{V\}, \lambda = \{1.0\}, v_{k+1} = V$\;
	
	\Return $[W_k, \lambda, v_{k+1}]$ \;
\end{algorithm}

For the version of implementation of our algorithms that tracks the order of the simplex vertex in the minimum simplex set $\tau_k$, the cases of S2D can be reduced to 4 instead of 7. For example, if the simplex vertex A is added most recently, the possible value of the barycode reduces to 4, 5, 6, 7.
This can be proved easily. Since A is the vertex added recently, the B, C form a hyperplane orthogonal to the supporting direction $-v_k$. The origin clearly lies in the $-v_k$ direction side of the BC hyperplane according to the definition. The barycode region B, C cone regions and the BC edge region lie in the $v_k$ direction side of the hyperplane BC, which is the other side of the hyperplane. Thus the barycode 1, 2, 3 cases are eliminated. For these implementations that do not track the order of the simplices, the Alg. \ref{alg:s2d} provides a general solution. Combining with Eq. \eqref{alg:s2d}, our S2D subroutine can be used to resolve point-in-triangle detection problem or integrated to 3D SV-based GJK algorithm to improve efficiency further. 

\noindent \textbf{S1D subroutine} For the S1D problem, there are only three possible voronoi regions: the two vertex and one edge regions. This is can be solved trivially by leveraging the orthogonal conditions (line 1, 3 in Alg. \ref{alg:s1d}) and invariance of the barycentric coordinates under affine transformations, shown in Alg. \ref{alg:s1d}.

\begin{algorithm} \label{alg:s1d}
	\caption{S1D($\tau_k$)}
	\KwIn{$\tau_k = \{ A, B\}$ the new simplex set}
	\KwOut{\\
		$W_k$ the minimal simplex set,\\
		$\lambda$ barycentric coordinats,\\
		$v_{k+1}$ the minimum distance vector}
	\SetKwFunction{S}{S}
	\SetKwFunction{SubDistance}{SubDistance}
	%	$\overrightarrow{OA}$\;
	\If{$\overrightarrow{OA} \cdot \overrightarrow{AB} \geq 0$}{
		$W_k = \{A\}, \lambda = \{1.0\}, v_{k+1} = A, \tau_k = W_k$\;
		\Return [$W_k, \lambda, v_{k+1}$]
%		\Return S0D$(\tau_k = \{A\})$\;
	}
	\If{$\overrightarrow{OB} \cdot \overrightarrow{AB} \leq 0$}{
		$W_k = \{B\}, \lambda = \{1.0\}, v_{k+1} = B, \tau_k = W_k$\;
		\Return [$W_k, \lambda, v_{k+1}$]\;
	}
	$sum = \overrightarrow{OA} \cdot \overrightarrow{AB} - \overrightarrow{OB} \cdot \overrightarrow{AB} $\;
	$\lambda_u$ = $\frac{- \overrightarrow{OB} \cdot \overrightarrow{AB}}{sum}, \lambda_v$ = $\frac{\overrightarrow{OA} \cdot \overrightarrow{AB}}{sum}, \lambda = \{ \lambda_u, \lambda_v\}$\;
	$W_k = \tau_k, v_{k+1} = \lambda_u \cdot A + \lambda_v \cdot B$\;
	\Return $[W_k, \lambda, v_{k+1}]$\;
\end{algorithm}

%Although the GJK algorithm is a general and efficient collision detection algorithm, it can't be used for calculating the penetration depth. The expanding polytope algorithm (EPA) is such an algorithm that is able to resolve the aforementioned problem. Even better, the minimum simplex $W_k$ of the GJK, can be used by the EPA as a warm start.
The proposed algorithms above are designed to solve the Level 2 distance query problem. Although we did not present pseudo codes related to the witness points extraction, it's trivial to implement it in our framework and the underlying cost is negligible. In our following benchmarks, the distance query time of our method actually includes the witness point extraction part.

\section{Collision Detection Sub-Routine}

\begin{figure}
	\centering
	\subfigure[seperating hyperplane for distant cases]{
		\label{fig:hyperplane} %% label for first subfigure
		\includegraphics[width=1.4in]{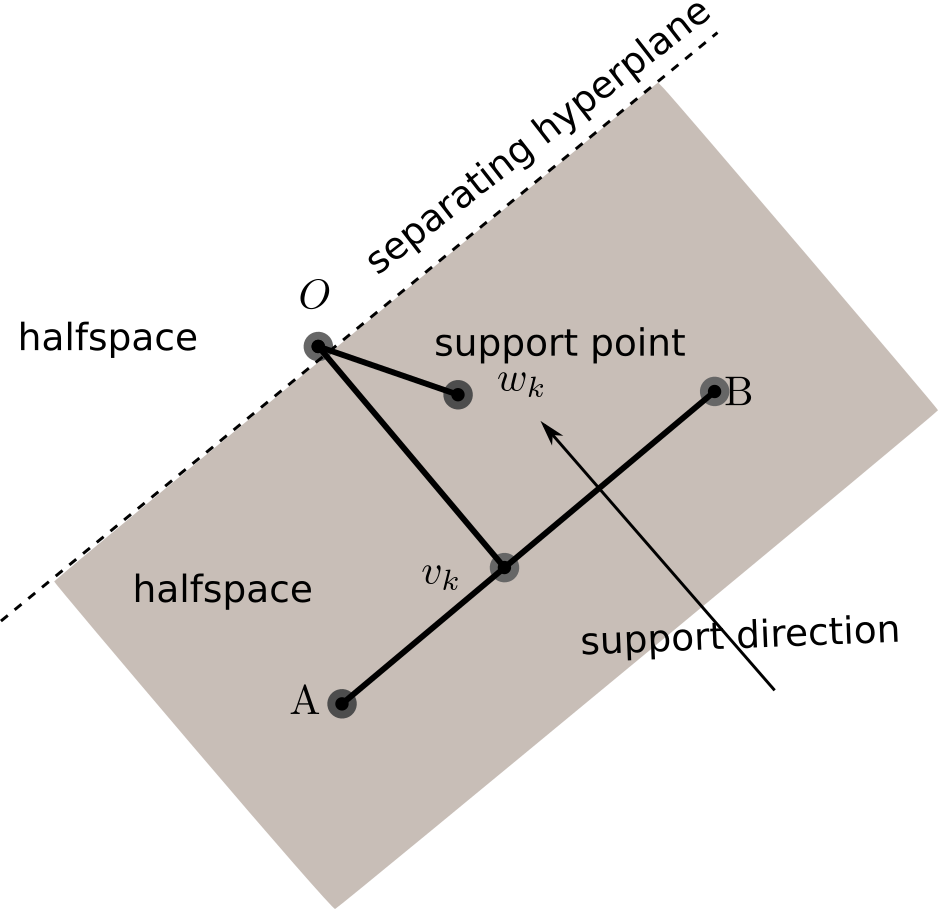}
	}
	\hspace{0.05in}
	\subfigure[early exit condition for overlap cases]{
		\label{fig:overlap_region} %% label for second subfigure
		\includegraphics[width=1.4in]{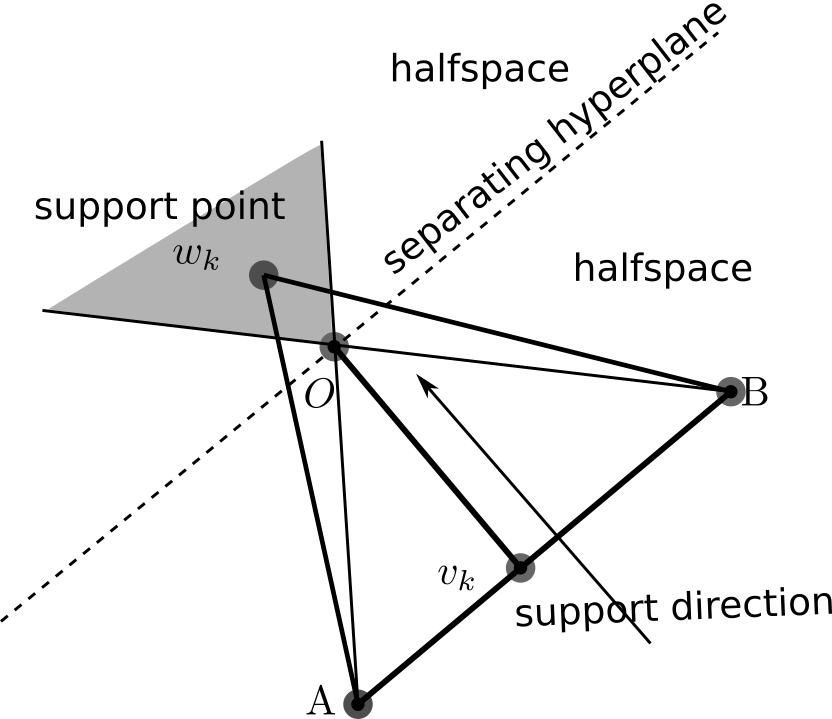}
	}
	\caption{early exit conditions for Level 1 collision detection}
	\label{fig:quickexit} %% label for entire figure
\end{figure}

\begin{algorithm} \label{alg:bgjk}
	\caption{GJK Collision Detection Algorithm}
	\KwIn{Convex  Shape:${P, Q}$}
	\SetKwFunction{false}{false}
	\SetKwFunction{true}{true}
	\KwOut{\true for collision, otherwise \false}
	\SetKwFunction{S}{S}
	\SetKwFunction{subdistance}{subdistance}
	\SetKwFunction{card}{card}
	\Operation{\card cardinality of a set}
	\Parameter{$k_{\text{max}}$ max iteration number allowed}
	\Parameter{$\varepsilon \in R_{++}$ the positive tolerance}
	\If{$\big(\exists v \in (\text{P} \ominus \text{Q})\big) \land (\lVert v \rVert \neq 0)$}
	{
		$v_{\text{init}}$ = $v$\;
	}
	\Else{
		\Return \true;
	}
	k = 0,
	$v_1$ = $\S_{\text{P} \ominus \text{Q}}(-v_{\text{init}})$\;
	$\tau_k = \{v_1\}$, $\text{W}_k = \{v_1\}$ \;
	\Do{ $ \big(\card(W_k) < 3 \big) \land (k < k_{\text{max}} )$}
	{
		k = k+1,
		$w_k$ = $\S_{\text{P} \ominus \text{Q}}(-v_k)$\;
		\If{$v_k \cdot w_k > 0$} {
			\Return \false\;
		}
		\Else{
			\If{\card{$W_k$} == 2} {
				$\{ A, B \} = W_k$\;
				\If{($A \times w_k$) $\cdot$ ($B \times w_k$) $ \leq 0$}{
					\Return \true;
				}
			}
			
		}
		$\tau_k = {w_k} \cup W_{k-1}$\;
		[$W_k, \lambda, v_{k+1}$] = \subdistance($\tau_k$)\;
		\If{$\lVert v_{k+1} \rVert \leq \varepsilon $} {
			\Return \true\;
		}
	}
	\Return ($\lVert v_{k+1} \rVert < \varepsilon \ ? \ \true : \false$) \;
\end{algorithm} 
According to the requirement level for collision detection defined in the problem formulation part, not all the operations in our algorithm are necessary for applications with a low level requirement. For example, the witness points and distance scalar are not needed for the Level 1 collision detection. Thus we present a specialized collision detection sub-routine that is able to tremendously improve the detection efficiency with only minor modifications on the exit conditions of the GJK. The modifications base on two facts. For distant collision detection cases, as long as the the lower boundary of the distance is positive, there always exists a separating hyperplane between the origin $O$ and the CSO polygon. $w_k$ is the furthest point of CSO in $-v_k$ direction. The $v_k \cdot w_k > 0$ condition at line 5 of the Alg. \ref{alg:bgjk} indicates $w_k$ is not further than the hyperplane that is going through the origin while perpendicular to the last distance vector $v_k$. In fact, all the CSO vertices are in the halfspace that includes $v_k$. Since the origin $O$ is in the other closed half space, $O \cap \text{CSO} = \emptyset$. In this case, there is no collision and the algorithm terminates, shown as Fig. \ref{fig:hyperplane}. It also needs less iterations and computation resources to find this separating hyperplane than the exact final $v_k$. Clearly, this condition is able to filter out a large portion of distant cases, but not all. For example, for a support point $w_k$ that is in the same half space with the origin O, it's possible that there is still no collision at all. This is a common speedup condition used by \cite{bergen1999fast, cameron1997enhancing} and many other implementations.

We go further by analysing the cases that $w$ goes beyond the hyperplane. In fact, when the $w_k$ is further than the hyperplane through the origin and falls into the region defined by the vertical angle region of $\angle{AOB}$ (shown as Fig. \ref{fig:overlap_region}), the triangle consisted of $A, B, w_k$ encloses the origin $O$ for sure. And this situation can be checked by a cheaper condition comparing to the barycode calculation. Thus there is no need to go through the sub-distance calculation process. This condition improves the efficiency for overlap cases further.

%\subsection{EPA}

%\section{GJK Optimization}
%\subsection{Local Support}
%\subsection{Data Type}
%\subsection{Hill Climbing}
%\subsection{With Motions}

\section{Results and Benchmarks}
In order to demonstrate the efficiency of our new GJK algorithm and its variants, we implement a benchmark dataset generator to randomly generate convex shapes (P, Q) with different vertex numbers (4, 8, 12, 16, 20, 24). To cover all the use cases of different applications for collision detection, we divide the dataset to three types: distant, touching and overlap cases according to the collision status between the P and Q shapes.
Then we apply proper random SE2 transformations (1000 random cases per type) to above shapes according to the corresponding collision status to get the final datasets. The random distant and overlap case datasets are trivial to get. The touching case dataset need a little bit more extra work. We start with the distant shape pairs and call the GJK algorithm of the Box2D to get the separating distance vector. Then by shifting one of the shapes according to the separating distance vectors, we generate the touching case dataset. 

\begin{figure}[htbp]
\centering

\subfigure[Level 1 collision detection]{
	\label{fig:binary_bench}
	\begin{minipage}[b]{0.5\linewidth}
		\centering
		\includegraphics[width=1\linewidth]{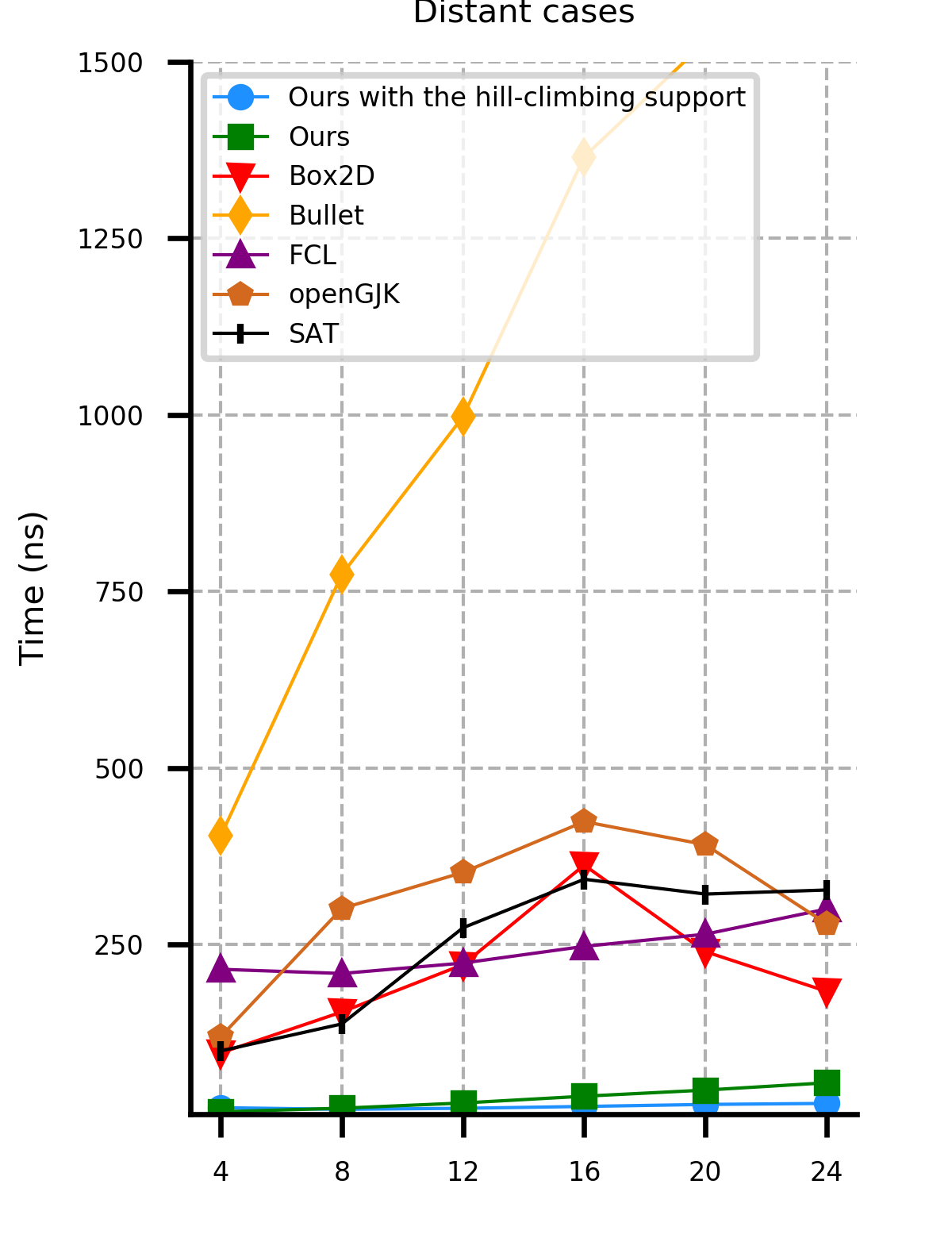}\vspace{1pt}
		\includegraphics[width=1\linewidth]{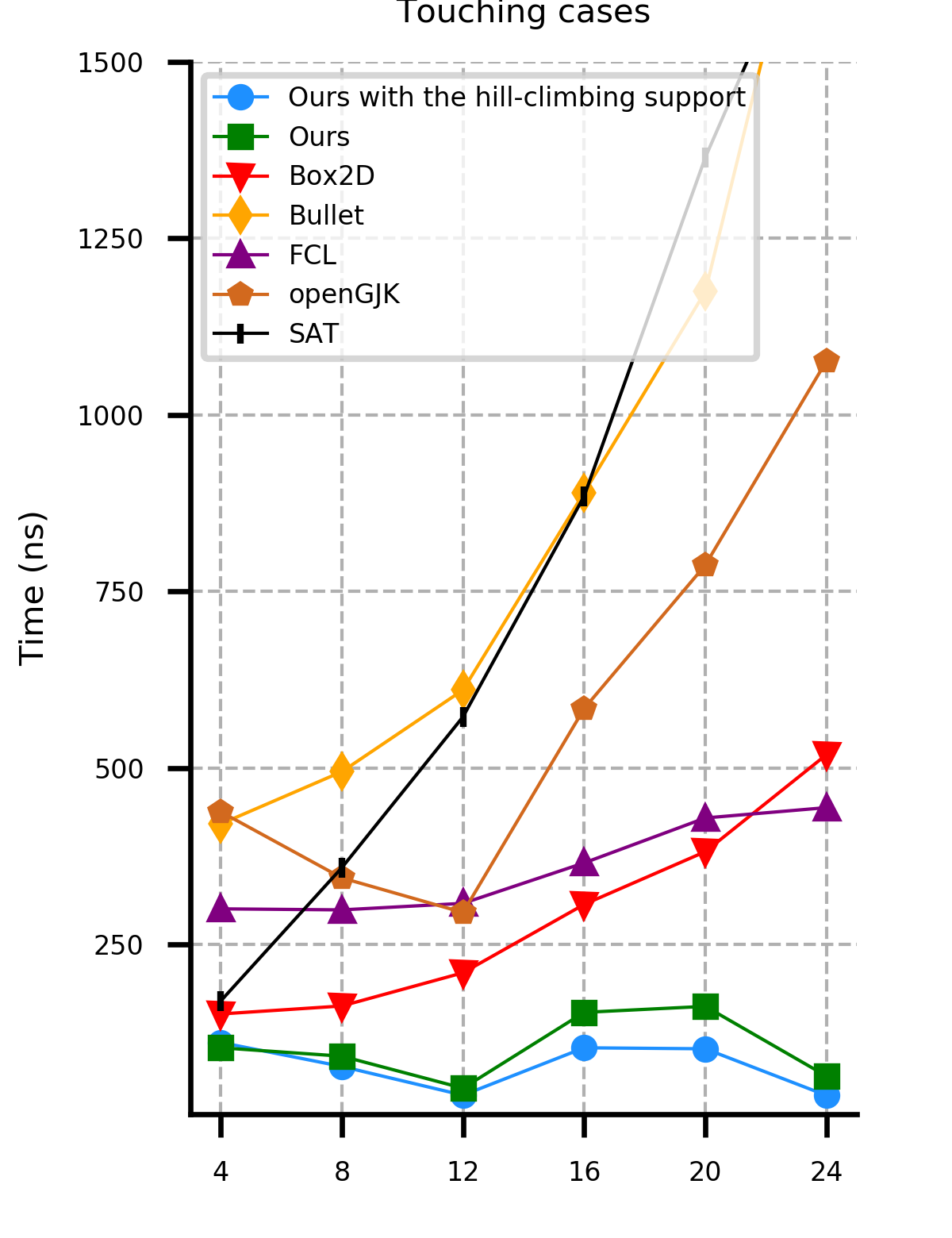}\vspace{1pt}
		\includegraphics[width=1\linewidth]{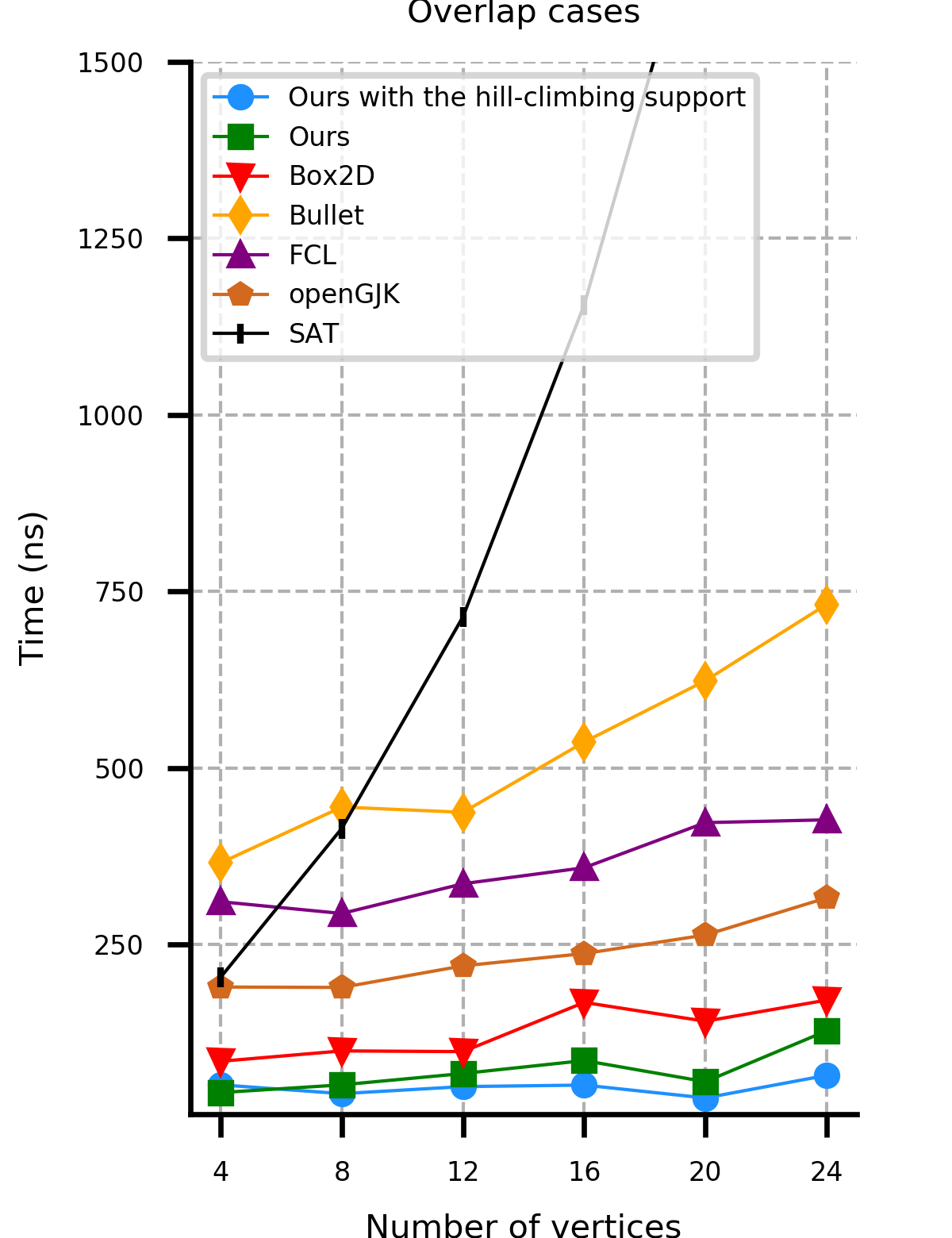}\vspace{1pt}
		%\caption{fig1}
	\end{minipage}%
}%
\subfigure[Level 2 distance query]{
	\label{fig:distance_bench}
	\begin{minipage}[b]{0.5\linewidth}
		\centering
		\includegraphics[width=1\linewidth]{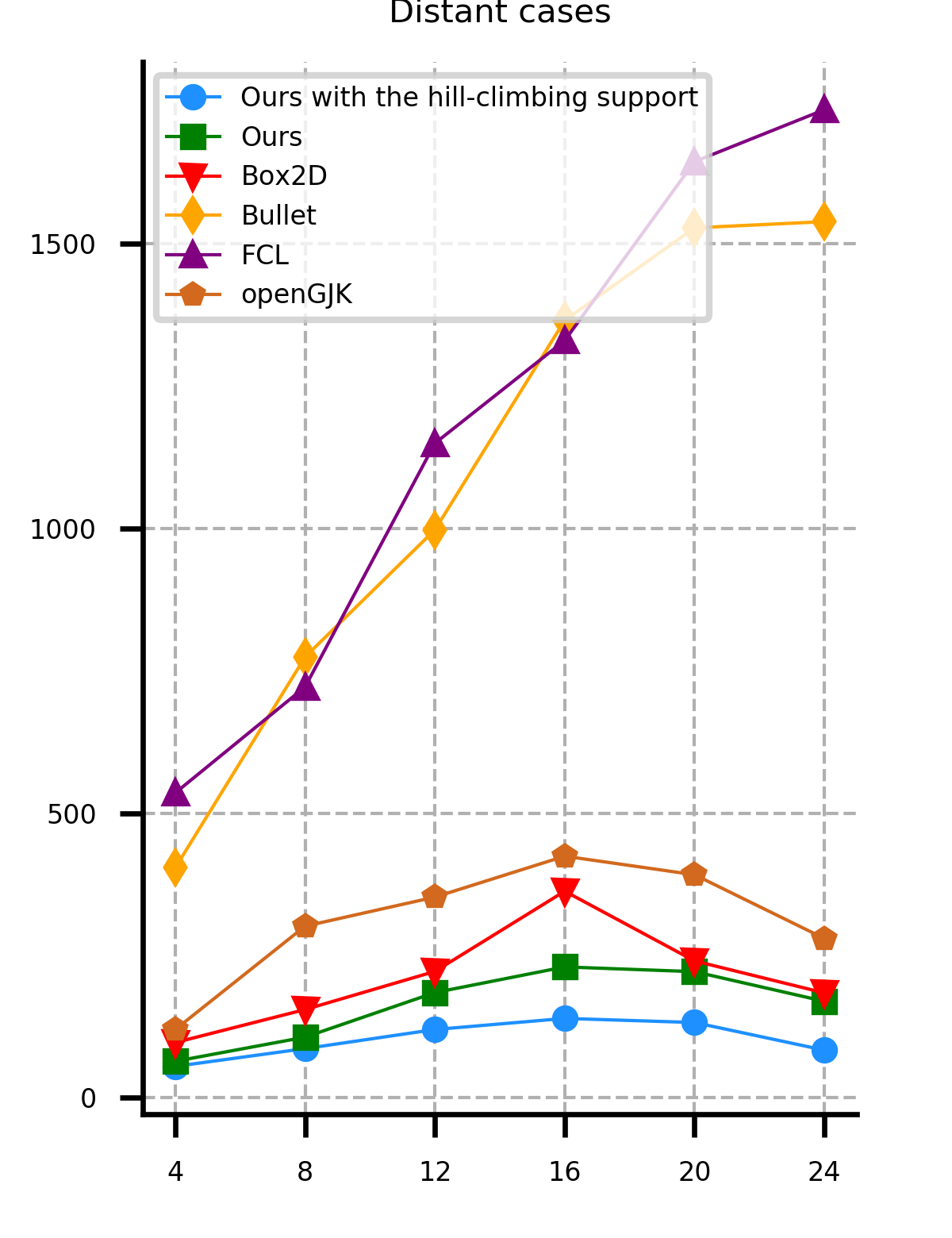}\vspace{1pt}
		\includegraphics[width=1\linewidth]{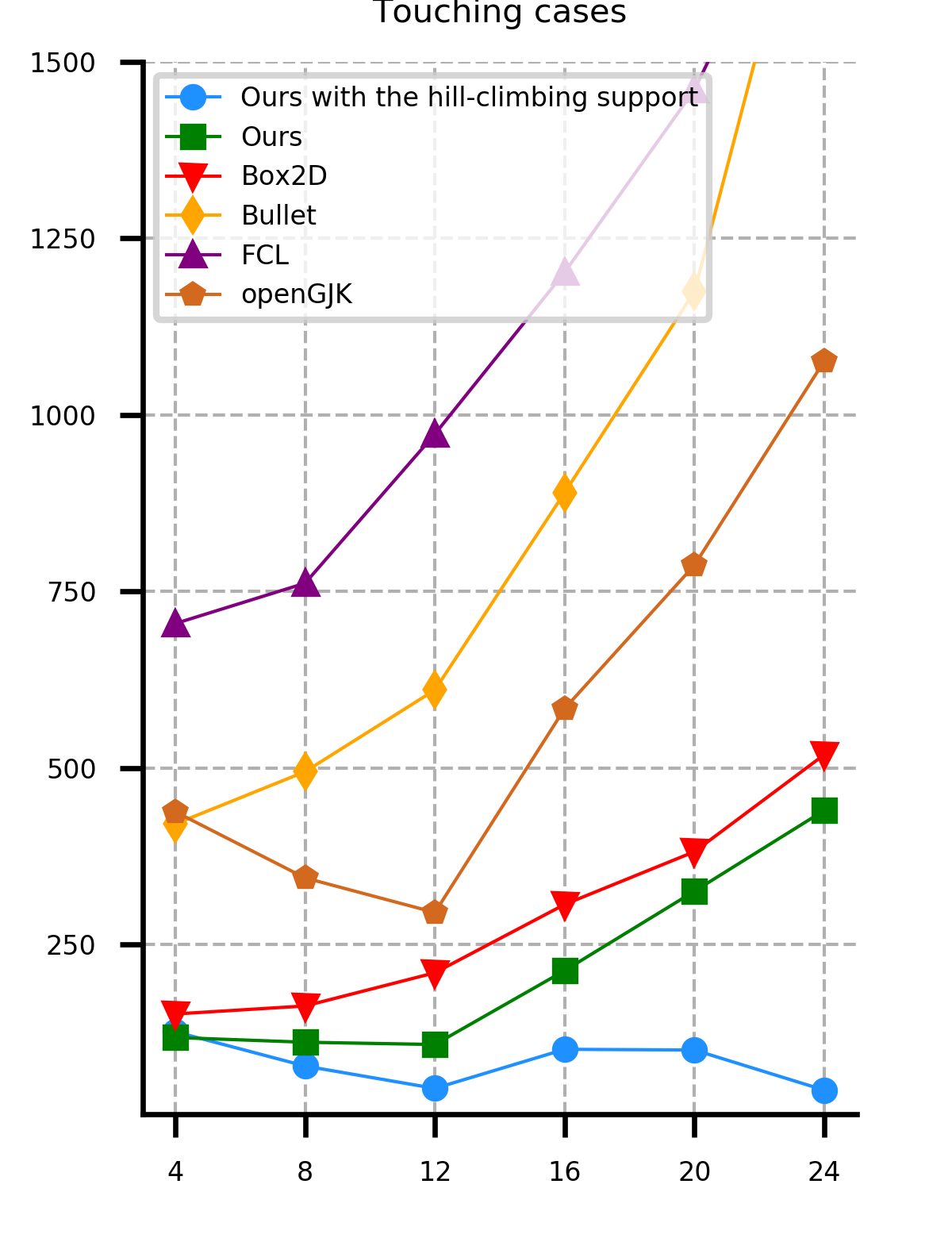}\vspace{1pt}
		\includegraphics[width=1\linewidth]{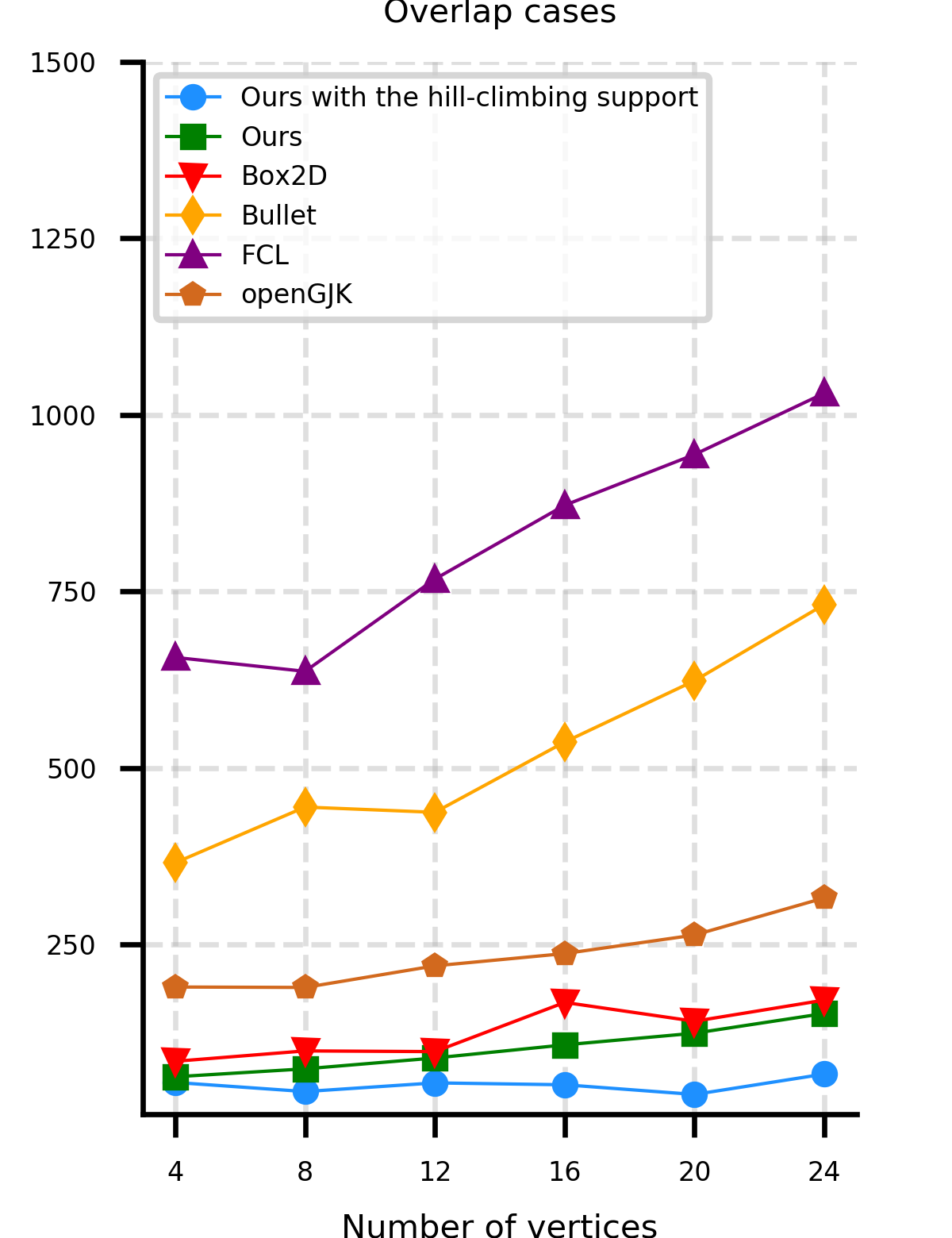}\vspace{1pt}
		%\caption{fig1}
	\end{minipage}%
}%

%	\subfigure[collision:distant]{
%		\begin{minipage}[b]{0.5\linewidth}
%			\centering
%			\includegraphics[width=1.2in]{images/collision_distant.png}\vspace{4pt}
%			\label{fig:simple_parallel_extremes}
%		\end{minipage}%
%	}%
%	\subfigure[distance:distant]{
%		\begin{minipage}[b]{0.5\linewidth}
%			\centering
%			\includegraphics[width=1.2in]{images/distance_distant.png}
%			%\caption{fig2}
%			\label{fig:common_multiple_extremes}
%		\end{minipage}%
%	}%
%	
%	\subfigure[collision:touching]{
%		\begin{minipage}[b]{0.5\linewidth}
%			\centering
%			\includegraphics[width=1.2in]{images/collision_touching.png}
%			%\caption{fig2}
%			\label{fig:mextremes}
%		\end{minipage}
%	}%
%	\subfigure[distance:touching]{
%		\begin{minipage}[b]{0.5\linewidth}
%			\centering
%			\includegraphics[width=1.2in]{images/distance_touching.png}
%			%\caption{fig2}
%			\label{fig:colinear_simplices}
%		\end{minipage}
%	}%
%	
%	\subfigure[collision:overlap]{
%		\begin{minipage}[b]{0.5\linewidth}
%			\centering
%			\includegraphics[width=1.2in]{images/collision_overlap.png}
%			%\caption{fig2}
%			\label{fig:mextremes}
%		\end{minipage}
%	}%
%	\subfigure[distance:overlap]{
%		\begin{minipage}[b]{0.5\linewidth}
%			\centering
%			\includegraphics[width=1.2in]{images/distance_overlap.png}
%			%\caption{fig2}
%			\label{fig:colinear_simplices}
%		\end{minipage}
%	}%
	
\centering
\caption{GJK benchmarks}
\vspace{-0.02cm}
\label{fig:gjkbench}
\end{figure}

According to the requirements level for collision detection, we do two groups of benchmarks over above three types of random datasets: Level 1 collision detection benchmark, and Level 2 distance query benchmark. For Level 1 collision detection, we only care about whether it is collision-free or not. We compared our binary GJK (Alg. \ref{alg:bgjk}) implementation with that of Bullet (3D), FCL (3D), OpenGJK (3D), Box2D (2D). Among them, only FCL provides a 3D implementation of the binary GJK algorithm. For other libraries, we use their distance interfaces for Level 1 benchmark in this group. Besides, we also benchmark the separate axis theorem (SAT, 2D) algorithm in this group for comparison. For all the three cases, our binary version (green lines in Fig. \ref{fig:binary_bench}) performs much better than all the open-source implementations in terms of efficiency, including the most efficient open-source version -- Box2D.

For Level 2 distance query benchmark, we compare our distance GJK implementation (Alg. \ref{alg:gjk}) with that of Bullet, FCL, OpenGJK and Box2D. Since the SAT algorithm can't provide precise distance information, it's not considered in this benchmark group. Among them, Bullet (3D), FCL (3D) and Box2D  (2D) use voronoi-based GJK implementations and OpenGJK (3D) employs SV-based GJK implementation. All the implementations are able to provide the right distance scalar. The results indicate 2D specialized GJK implementations provides remarkable improvements over 3D versions in efficiency. Our new GJK distance algorithm (green lines in Fig. \ref{fig:distance_bench}) is the most efficient algorithm in all random distant, touching and overlap cases. We also benchmark the distance implementation of polygon2d in Baidu Apollo. It takes hundreds of times more than ours, which is beyond the range of the figure. Thus we did not show their results in the Fig. \ref{fig:gjkbench}. We provide the detailed benchmark runtime data in the appendix for your reference.
%For all three cases, our algorithm shows about the same average calculation time while Box2D shows a clear time increase in random touching dataset. This demonstrates a consistent algorithm behavior of our method for different use cases.

Another point that could affect the efficiency of GJK algorithms is the start support direction. A bad support direction will increase the iteration number of the GJK algorithm. There actually two common-used ways to pick a support direction. The first one is to use a fixed direction for every collision detection. The second is to use an arbitrary point in CSO as the support direction and the first simplex vertex as well. According to our tests, for a random dataset of convex shapes with 4 vertices, the second way is able to reduce the GJK calculation time to the half on average. Ours and Box2D both use the second way to pick the start support direction.

Since vertices of all the convex shapes are required to to counter-clock-wise in 2D, it's trivial to enhance the support function with the hill-climbing strategy to speed up the GJK algorithms. Thus we also benchmark our algorithms with the hill-climbing support in these two benchmark groups. Due to the limited space, we do not present the exact pseudo codes of it here. It's trivial to implement in 2D without requirements for any special data structure. The corresponding results are shown as light blue lines in Fig. \ref{fig:gjkbench}, which currently is the best in all cases. With the hill-climbing strategy, we can improve the efficiency further, especially for the convex shapes with many vertices.

All the benchmarks are done with the help of the google benchmark library on a PC with an Intel Xeon E3 processor at 2.8GHz and 8GB RAM in a Linux system. The proposed algorithms have been applied to the autonomous driving system at UISEE Technology Co. Ltd..

% epa standalone algorithm vs gjk + epa
\section{CONCLUSIONS}

In this paper, we present a more efficient GJK algorithm to solve the collision detection and distance query problems in 2D.
First, we propose a new barycode-based GJK distance algorithm to improve the efficiency of distance query. Then, we optimize the exit conditions of our GJK distance algorithm to provide a highly efficient collision detection sub-routine, which shows dramatic improvements on run-time for applications that only care about binary results. We benchmark our methods along with that of the well-known open-source collision detection libraries, such as Bullet, FCL, OpenGJK, Box2D, and Apollo over a range of random datasets. The results indicate that our methods and implementations outperform the state-of-the-art in both collision detection and distance query.
%\section*{ACKNOWLEDGMENT}

\bibliographystyle{IEEEtran}
\bibliography{IEEEabrv,IEEEexample}

\section*{APPENDIX}
The tables show the average time (in nanosecond) of different algorithms for one collision detection on the random datasets that is related to Fig. \ref{fig:gjkbench}. HCS is short for the hill-climbing support. The Baidu Apollo is refered to its polygon2d distance implementation.

\begin{table}[!t]
	\caption{Average time for 4-vertices random datasets}
	\tiny  
	\label{table_time}
	\begin{tabular}{lllllll}  
		  \toprule  
		  \multirow{2}{*}{\textbf{Alg.}}&  
		  \multicolumn{3}{c}{\textbf{Distance (ns)}}&\multicolumn{3}{c}{\textbf{Collision (ns)}}\cr  
		  \cmidrule(lr){2-4} \cmidrule(lr){5-7}  
		  &\textbf{Distant} & \textbf{Overlapping} & \textbf{Touching} & \textbf{Distant}& \textbf{Overlapping} & \textbf{Touching} \cr  
		  \midrule   
		
	Box2D& 96.77& 84.93& 151.92& 96.77& 84.93& 151.92\\
	
	SAT& -& -& - & 99.62& 203.92& 170.26\\
	\textbf{Ours(without HCS)}& \textbf{63.85}& \textbf{62.99}& \textbf{118.61}& \textbf{13.45}& \textbf{40.63}& \textbf{104.05}\\
	
	\textbf{Ours(with HCS)}& \textbf{54.92}& \textbf{54.95}& \textbf{126.73}& \textbf{19.22}& \textbf{51.54}& \textbf{111.59}\\

	FCL& 535.41& 656.78& 704.75& 215.19& 310.85& 300.77\\

	Bullet& 403.87& 366.01& 420.92& 403.87& 366.01& 420.92\\
	
	OpenGJK& 119.66& 190.09& 438.25& 119.66& 190.09& 438.25\\
	
	Baidu Apollo& 2731.96& 549.41& 2586.54& -& -& -\\

		\bottomrule  
		
	\end{tabular}
	
\end{table}

\begin{table}[!t]  \caption{Average time for 8-vertices random datasets}

	\tiny  
	\label{table_time}
	
	\begin{tabular}{lllllll}  
		\toprule  
		\multirow{2}{*}{\textbf{Alg.}}&  
		\multicolumn{3}{c}{\textbf{Distance (ns)}}&\multicolumn{3}{c}{\textbf{Collision (ns)}}\cr  
		\cmidrule(lr){2-4} \cmidrule(lr){5-7}  
		&\textbf{Distant} & \textbf{Overlapping} & \textbf{Touching} & \textbf{Distant}& \textbf{Overlapping} & \textbf{Touching} \cr  
		\midrule 
		
		%		\toprule   
		%		
		%		cases & distant & overlapping & touching & distant & touching & overlapping \\  
		%		
		%		\midrule   
		
		Box2D& 154.77& 99.63& 163.24& 154.77& 99.63& 163.24\\
		
		SAT& -& -& -& 137.83& 414.83& 359.05\\
		
		\textbf{Ours(without HCS)}& \textbf{106.25}& \textbf{74.25}& \textbf{111.90}& \textbf{18.50}& \textbf{51.58}& \textbf{92.18}\\
		
		\textbf{Ours(with HCS)}& \textbf{86.13}& \textbf{42.07}& \textbf{78.21}& \textbf{16.90}& \textbf{39.13}& \textbf{77.63}\\

		FCL& 721.65& 637.04& 761.84& 209.25& 294.20& 299.31\\

		Bullet& 774.23& 444.89& 494.98& 774.23& 444.89& 494.98\\
		
		OpenGJK& 301.28& 189.47& 344.58& 301.28& 189.47& 344.58\\
		
		Baidu Apollo& 8593.92& 7059.60& 8381.86&  -& -& -\\
		
		\bottomrule  
		
	\end{tabular}
	
\end{table}

\begin{table}[!t]  \caption{Average time for 12-vertices random datasets}

	\tiny  
	\label{table_time}
	
	\begin{tabular}{lllllll}  
		\toprule  
		\multirow{2}{*}{\textbf{Alg.}}&  
		\multicolumn{3}{c}{\textbf{Distance (ns)}}&\multicolumn{3}{c}{\textbf{Collision (ns)}}\cr  
		\cmidrule(lr){2-4} \cmidrule(lr){5-7}  
		&\textbf{Distant} & \textbf{Overlapping} & \textbf{Touching} & \textbf{Distant}& \textbf{Overlapping} & \textbf{Touching} \cr  
		\midrule  
%		Box2d& 221.77& 98.62& 210.26& 221.77& 98.62& 210.26\\
%		
%		SAT& -& -& -& 274.01& 714.44& 572.71\\
%		
%		ours& 184.58& 89.60& 108.69& 25.89& 67.62& 46.81\\
%		
%		FCL& 1149.75& 54.34& 46.52& 18.37& 49.10& 36.89\\
%		
%		Bullet& 997.69& 437.45& 610.69& 997.69& 437.45& 610.69\\
%		
%		OpenGJK& 352.38& 220.07& 295.60& 352.38& 220.07& 295.60\\
%		
%		Baidu Apollo& 17864.63& 16083.23& 23822.42& 17864.63& 16083.23& 23822.42\\
%		
		  
	Box2D& 221.77& 98.62& 210.26& 221.77& 98.62& 210.26\\
	
	SAT& -& -& -& 274.01& 714.44& 572.71\\
	
	\textbf{Ours(without HCS)}& \textbf{184.58}& \textbf{89.60}& \textbf{108.69}& \textbf{25.89}& \textbf{67.62}& \textbf{46.81}\\
	
	\textbf{Ours(with HCS)}& \textbf{119.69}& \textbf{54.34}& \textbf{46.52}& \textbf{18.37}& \textbf{49.10}& \textbf{36.89}\\

	FCL& 1149.75& 767.70& 973.09& 224.02& 336.34& 308.55\\

	Bullet& 997.69& 437.45& 610.69& 997.69& 437.45& 610.69\\
	
	OpenGJK& 352.38& 220.07& 295.60& 352.38& 220.07& 295.60\\
	
	Baidu Apollo& 17864.63& 16083.23& 23822.42&-& -& -\\
		
		\bottomrule  
		
	\end{tabular}
	
\end{table}

\begin{table}[!t]  \caption{Average time for 16-vertices random datasets}
	\tiny  
	\label{table_time}
	
	\begin{tabular}{lllllll}  
		\toprule  
		\multirow{2}{*}{\textbf{Alg.}}&  
		\multicolumn{3}{c}{\textbf{Distance (ns)}}&\multicolumn{3}{c}{\textbf{Collision (ns)}}\cr  
		\cmidrule(lr){2-4} \cmidrule(lr){5-7}  
		&\textbf{Distant} & \textbf{Overlapping} & \textbf{Touching} & \textbf{Distant}& \textbf{Overlapping} & \textbf{Touching} \cr  
		\midrule

	Box2D& 362.73& 168.27& 307.07& 362.73& 168.27& 307.07\\
	
	SAT& -& -& -& 342.67& 1156.10& 885.26\\
	
	\textbf{Ours(without HCS)}& \textbf{229.68}& \textbf{108.19}& \textbf{213.41}& \textbf{35.49}& \textbf{85.96}& \textbf{154.20}\\
	
	\textbf{Ours(with HCS)}& \textbf{139.18}& \textbf{51.67}& \textbf{101.88}& \textbf{20.98}& \textbf{51.23}& \textbf{104.03}\\
	
	FCL& 1330.37& 872.25& 1202.26& 247.63& 359.01& 365.84\\

	Bullet& 1365.69& 536.98& 890.28& 1365.69& 536.98& 890.28\\
	
	OpenGJK& 424.26& 237.44& 584.42& 424.26& 237.44& 584.42\\
	
	Baidu Apollo& 30070.27& 1973.60& 29700.82& -& -& -\\

		\bottomrule  
		
	\end{tabular}
	
\end{table}

\begin{table}[ht]  \caption{Average time for 20-vertices random datasets}
	\tiny  
	\label{table_time}
	
	\begin{tabular}{lllllll}  
		\toprule  
		\multirow{2}{*}{\textbf{Alg.}}&  
		\multicolumn{3}{c}{\textbf{Distance (ns)}}&\multicolumn{3}{c}{\textbf{Collision (ns)}}\cr  
		\cmidrule(lr){2-4} \cmidrule(lr){5-7}  
		&\textbf{Distant} & \textbf{Overlapping} & \textbf{Touching} & \textbf{Distant} & \textbf{Overlapping} & \textbf{Touching}\cr  
		\midrule 

		Box2D& 240.20& 141.63& 381.77& 240.20& 141.63& 381.77\\
		
		SAT& -& -&-& 321.73& 1755.92& 1364.87\\
		
		\textbf{Ours(without HCS)}& \textbf{221.33}& \textbf{124.68}& \textbf{325.81}& \textbf{44.15}& \textbf{56.04}& \textbf{162.70}\\
	
		\textbf{Ours(with HCS)}& \textbf{131.98}& \textbf{37.95}& \textbf{100.88}& \textbf{23.84}& \textbf{33.00}& \textbf{102.61}\\
		
		FCL& 1644.64& 944.10& 1460.72& 264.87& 422.83& 429.47\\

		Bullet& 1528.36& 623.79& 1174.94& 1528.36& 623.79& 1174.94\\
		
		OpenGJK& 391.63& 263.68& 787.58& 391.63& 263.68& 787.58\\
		
		Baidu Apollo& 45164.85& 11262.01& 44436.56& -& -& -\\
		
		\bottomrule  
		
	\end{tabular}
	
\end{table}

\begin{table}[]
	\caption{Average time for 24-vertices random datasets}
	\tiny  
	\label{table_time}
	
	\begin{tabular}{lllllll}  
		\toprule  
		\multirow{2}{*}{\textbf{Alg.}}&  
		\multicolumn{3}{c}{\textbf{Distance (ns)}}&\multicolumn{3}{c}{\textbf{Collision (ns)}}\cr  
		\cmidrule(lr){2-4} \cmidrule(lr){5-7}  
		&\textbf{Distant} & \textbf{Overlapping} & \textbf{Touching} & \textbf{Distant}& \textbf{Overlapping} & \textbf{Touching} \cr  
		\midrule 
		
		%		\toprule   
		%		
		%		cases & distant & overlapping & touching & distant & touching & overlapping \\  
		%		
		%		\midrule   
		
	Box2D& 183.96& 171.54& 519.16& 183.96& 171.54& 519.16\\
	
	SAT& -& -& -& 327.45& 2375.39& 1757.84\\
	
	\textbf{Ours(without HCS)}& \textbf{169.26}& \textbf{152.61}& \textbf{440.28}& \textbf{54.47}& \textbf{127.93}& \textbf{63.94}\\
	
	\textbf{Ours(with HCS)}& \textbf{83.41}& \textbf{66.83}& \textbf{44.02}& \textbf{25.25}& \textbf{64.97}& \textbf{36.56}\\
	
	FCL& 1735.86& 1030.91& 1841.25& 300.51& 426.83& 444.11\\
	
	Bullet& 1539.18& 731.75& 1876.92& 1539.18& 731.75& 1876.92\\
	
	OpenGJK& 279.72& 316.26& 1076.67& 279.72& 316.26& 1076.67\\
	
	Baidu Apollo& 64492.50& 8090.15& 64340.11& -& -& -\\
				
		\bottomrule  
		
	\end{tabular}
	
\end{table}

%
%
%% epa standalone algorithm vs gjk + epa
%\section{CONCLUSIONS}
%
%In this paper, we present a more efficient GJK algorithm to solve the collision detection and distance query problems in 2D.
%First, we propose a new barycode-based GJK distance algorithm to improve the efficiency of distance query. Then, we optimize the exit conditions of our GJK distance algorithm, which shows dramatic improvements on run-time for applications that only care about binary results. We benchmark our methods along with that of the well-known open-source collision detection libraries, such as Bullet, FCL, OpenGJK, Box2D, and Apollo over a range of random datasets. The results indicate that our methods and implementations outperform the state-of-the-art in both collision detection and distance query.
%\section*{ACKNOWLEDGMENT}
%
%\bibliographystyle{IEEEtran}
%\bibliography{IEEEabrv,IEEEexample}

\end{document}